\documentclass[conference]{IEEEtran}
\IEEEoverridecommandlockouts
\usepackage{cite}
\usepackage{float}
\usepackage{caption}
\usepackage{booktabs} 
\usepackage[caption=false,font=footnotesize]{subfig}
\usepackage{amsmath,amssymb,amsfonts}
\usepackage{algorithmic}
\usepackage{graphicx}
\usepackage{textcomp}
\usepackage{xcolor}
\def\BibTeX{{\rm B\kern-.05em{\sc i\kern-.025em b}\kern-.08em
    T\kern-.1667em\lower.7ex\hbox{E}\kern-.125emX}}
\begin{document}

\title{Corner Reflector Array Jamming Discrimination Using Multi-Dimensional Micro-Motion Features with Frequency Agile Radar

\thanks{This work was supported by the National Natural Science Foundation of China under Grants 62301295 and the corresponding author is Lei Wang (e-mail: leiwangqh@tsinghua.edu.cn).}
}

\author{\IEEEauthorblockN{1\textsuperscript{st}Jie Yuan, 2\textsuperscript{nd}Lei Wang*, 3\textsuperscript{rd}Yanhao Wang , 4\textsuperscript{th}Yimin Liu}
\IEEEauthorblockA{\textit{Department of Electronic Engineering}, \textit{Tsinghua University}\\
Beijing, China }}

\maketitle

\begin{abstract}
This paper introduces a robust discrimination method for distinguishing real ship targets from corner-reflector-array jamming with frequency-agile radar. The key idea is to exploit the multidimensional micro-motion signatures that separate rigid ships from non-rigid decoys. From Range–Velocity maps we derive two new hand-crafted descriptors—mean weighted residual (MWR) and complementary contrast factor (CCF)—and fuse them with deep features learned by a lightweight CNN. An XGBoost classifier then gives the final decision. Extensive simulations show that the hybrid feature set consistently outperforms state-of-the-art alternatives, confirming the superiority of the proposed approach.
\end{abstract}


\section{Introduction}
Effectively countering various passive and active interferences is a key capability of radar systems\cite{b1,b2}. Corner reflectors represent a typical type of passive interference false target in radar electronic countermeasure environments. Leveraging their metallic composition and unique geometric structure, corner reflectors can achieve a radar cross section (RCS) comparable to that of a large ship, despite their small physical size. Multiple corner reflectors can be arranged into an array deployed around a vessel, simulating ship targets through various configurations, thereby effectively deceiving anti-ship guidance radar detection and tracking systems. Consequently, analyzing the differences in radar echoes between ships and corner reflector arrays, and utilizing these distinctions for target identification, is of significant research importance. 

A unique correspondence exists between a target's micro-motion features and its structural and kinematic properties. Based on this principle, ships and corner reflectors exhibit notable structural differences and demonstrate distinct kinematic behaviors under identical sea conditions, allowing for their discrimination using micro-motion characteristics. Reference \cite{b2} investigated the motion model of sea-surface corner reflectors and the extraction of micro-Doppler features, demonstrating that micro-Doppler can be regarded as a unique signature of the target that provides additional information for target recognition \cite{b3,b4}. Meanwhile, reference \cite{b5} proposed a method for extracting relevant time-frequency characteristic parameters from the time-frequency representation of target echoes to distinguish between corner reflector jamming and ships, where reliable micro-Doppler signature classification requires the use of robust features that are capable of uniquely describing the micromotion \cite{b6}. However, current research on discriminating ship targets from corner reflector jamming based on micro-motion features predominantly focuses on the impact of single corner reflectors, leaving a notable gap in mature methodologies for dealing with corner reflector array jamming.

This paper proposes a recognition method for corner reflector array jamming based on multi-dimensional micro-motion features. First, we establish a target echo signal model for frequency-agile radar\cite{b7} and generate Range-Velocity maps(RV Maps) of the target through signal processing. By analyzing the differences in micro-motion characteristics between corner reflector arrays and ship targets, we design and extract two key discriminative features: the mean weighted residual (MWR) and the complementary contrast factor (CCF). These handcrafted features are subsequently combined with deep representations automatically extracted by a convolutional neural network (CNN) to form a multi-dimensional fused feature vector. Finally, an XGBoost classifier is employed to achieve high-accuracy identification of ships versus corner reflector arrays.Extensive comparative simulation experiments demonstrate the effectiveness of the proposed method.The method proposed in \cite{b8}, which addresses the ballistic target recognition using multi-dimensional micro-motion features in frequency-agile radars, provides an important reference for this work.

\section{Frequency-Agile Radar Echo Model and Target RV Map Generation}

\subsection{Frequency-agile Radar Echo Signal Model}

To obtain high resolution for each pulse in frequency-agile mode and support subsequent synthetic wideband processing, we adopt the Linear Frequency Modulated (LFM) signal as the baseband waveform. The signal model for the $n$-th pulse can be expressed as:
\begin{equation}
s_n(t)={\rm rect}\left(\frac{\hat{t}}{T_p}\right) \exp\left[{j2\pi \left(f_n\hat t+\frac{1}{2}\mu \hat{t}^2\right)}\right]\label{s(t)},
\end{equation}
where we define $\hat{t}=t-t_n$ as  fast time, $t_n=nT_r$ is slow time, $n=0,1,...,N-1$, $N$ is the number of coherent processing pulses, $T_r$ is the pulse repetition interval (PRI), $T_p$ is the pulse width, $\rm rect(t)$ equals 1 when $0 \leq t \leq 1$ and 0 otherwise, $\mu=B/T_p$ is the chirping rate and $B$ is bandwith, $\displaystyle f_n=f_0+C_n\Delta f$ is the carrier frequency for the $n$-th pulse, $f_0$ is the radar's initial carrier frequency, $C_n$ is the random frequency-hopping code, which is a random integer from the set$\{0,1,...,M-1\}$, $M$ is the number of distinct frequency points, and $\Delta f$ is the frequency step. Considering the target as a composition of scattering centers, the final received target echo can be expressed as the linear superposition of echoes from all scattering centers:
\begin{equation}
\begin{split}
    r_{n}(t)&=\sum\limits_{k=1}^{K}\sigma_k {\rm rect}\left(\frac{\hat t-\tau_{n,k}}{T_p}\right)\cdot \\
    &\exp{\left\{j2\pi\left[f_n(\hat t-\tau_{n,k})+\frac{1}{2}\mu (\hat t-\tau_{n,k})^2\right]\right\}},
\end{split}
\end{equation}
where $\sigma_k$ and $\tau_{n,k}$ are the  scattering amplitude and the time delay of the $k$-th scattering center, respectively, and $\tau_{n,k}$ is defined as $\tau_{n,k}={2(R_k+v_knT_r)}/{c}$. $R_k$ and $v_k$ are the initial radial distance and radial velocity, respectively, between the $k$-th scattering center on the target and the radar\cite{b9,b10}, and $c$ is the speed of light. The received echo signal is down-converted to obtain:
\begin{equation}
\begin{split}
\hat{r}_n(t)
&=\sum\limits_{k=1}^{K}\sigma_k{\rm rect}\!\left(\frac{\hat t-\tau_{n,k}}{T_p}\right)\cdot\\
&\exp\!\left\{j2\pi\!\left[-f_n\tau_{n,k}
+\frac{1}{2}\mu(\hat t-\tau_{n,k})^2\right]\right\} .
\end{split}
\end{equation}

\subsection{Echo Signal Processing and Target RV Map Generation}
According to (\ref{s(t)}), the frequency-domain representation of the transmitted signal $s_n(t)$ can be obtained as:

\begin{equation}
\begin{split}
S_n(f)=\sqrt{\frac{1}{\mu}}{\mathrm{rect}}\left(\frac{f}{B}\right)\cdot
\exp\left(-j\pi\frac{f^2}{2\mu}\right).
\end{split}
\end{equation}
Therefore, the frequency-domain representation of $\hat{r}_n(t)$ can be expressed as:

\begin{equation}
\begin{split}
\hat{R}_n(f)=\sum\limits_{k=1}^{K}\sigma_k\sqrt{\frac{1}{\mu}}{\rm rect}\left(\frac{f}{B}\right)
    \exp\!\left[-j2\pi\!\left(\frac{f^2}{2\mu}+f_n \tau_{n,k}\right)\right]\hspace{2cm}
    \label{eq5}.
\end{split}
\end{equation}
After pulse compression, the frequency-domain representation of the signal is obtained as:

\begin{equation}
    \begin{split}
        R_{\mathrm{pc}}(f)=\sum\limits_{k=1}^{K}\sigma_k\frac{1}{\mu}{\rm rect}\left(\frac{f}{B}\right)\exp(-j2\pi f_n \tau_{n,k}).
    \end{split}
\end{equation}
After Doppler compensation, the frequency-domain representation of the signal is:
\begin{equation}
    \begin{split}
       & R_{\mathrm{comp}}(f)=R_{\mathrm{pc}}(f)\cdot \exp{\left(-j2\pi\frac{2vf_nf}{c\mu}\right)}\\
                   &=\sum\limits_{k=1}^{K}\frac{\sigma_k}{\mu}{\rm rect}\left(\frac{f}{B}\right)\exp\left[-j2\pi f_n \left(\frac{2vf}{c\mu}+ \tau_{n,k}\right)\right].
    \end{split}
\end{equation}
Considering range cell migration correction, the frequency-domain expression of the compensated signal is:
\begin{equation}
    \begin{split}
       & R_{\mathrm{rmc}}(f)=R_{\mathrm{comp}}(f)\cdot \exp{\left(-j2\pi f_n\frac{2vnT_r}{c}\right)}\\
                   &=\sum\limits_{k=1}^{K}\frac{\sigma_k}{\mu}{\rm rect}\left(\frac{f}{B}\right)\cdot
                  \exp\left\{-j2\pi f_n \left[\frac{2v}{c}\left(\frac{f}{\mu}+nT_r\right)+ \tau_{n,k}\right]\right\}.
    \end{split}
\end{equation}

Since the carrier frequency of each pulse varies and is typically non-continuous, each spectral segment needs to be resampled onto a corresponding unified frequency axis. According to \cite{b11}, by accumulating signals in phase, we can obtain $S(f_i)=\frac{1}{N_i}\Sigma_{n_i=1}^{N_i}R_{\mathrm{rmc},n_i}(f_i)$, where $R_{rmc,n_i}(f_i)$ is the spectrum of the echo signal from the $n_i$-th pulse using carrier frequency $f_i$ after Doppler compensation and range migration correction, and $N_i$ denotes the number of pulses using carrier frequency $f_i$.Subsequently, the coherently integrated signal spectrum at each frequency point is resampled onto the synthesized frequency axis to obtain the complete target echo signal spectrum within one CPI. By performing an IFFT on this spectrum, the target's high-resolution range profile(HRRP) can be obtained.The total bandwidth is $B_{synth}=M\cdot\Delta f$, and the range resolution is $\Delta R=c/2B_{synth}$.

Since the radial velocities of individual scattering points on the target vary,we set multiple velocity grid points $v_i$ and perform Doppler compensation and range migration compensation on the target echoes according to the value of each grid point.This allows us to extract the radial velocities of the scattering points on the target through peak matching. Each $v_i$ corresponds to a ${\rm HRRP}_i$\cite{b12}.By concatenating this series of ${\rm HRRP}_i$ along the dimension of the predefined velocity grid points, a RV map of the target can be generated:  
${\rm RV=|Joint(HRRP}_i)|$, where “$\rm Joint$" means concatenating the HRRP generated at each grid point along the velocity grid dimension. The velocity resolution is: $\Delta v={\lambda}/{2NT_r}$. This map explicitly represents the coupled distance and velocity information of each scattering center on the target as detected by the radar.

\section{Recognition Method Design for Maritime Targets}

\subsection{Maritime Target Micro-motion Modeling and Characteristic Analysis}\label{AA}

Micro-motion refers to the small-amplitude, high-frequency, or quasi-periodic self-motions of a target relative to its main translational motion. The micro-motion of maritime targets is induced by wave-driven disturbing forces that push the target into movements in different directions.In three-dimensional space, it encompasses six degrees of freedom of motion, including three translational degrees of freedom—surge, sway, and heave—and three rotational degrees of freedom—roll, pitch, and yaw.These variables describe the translational motions of the target along its longitudinal, transverse, and vertical axes (surge, sway, and heave, respectively), as well as the rotational oscillations around its longitudinal, transverse, and vertical axes (roll, pitch, and yaw, respectively).Among these, the three rotational degrees of freedom are directly driven by wave-induced torques and can generate significant micro-Doppler effects, while the influence of the translational degrees of freedom is relatively uniform and their impact is less pronounced.Therefore, the micro-motion of maritime targets can be simplified to a model comprising only the three rotational micro-motions(roll, pitch, and yaw)\cite{b13}.

\begin{figure}[t]
\centerline{\includegraphics[width=6cm]{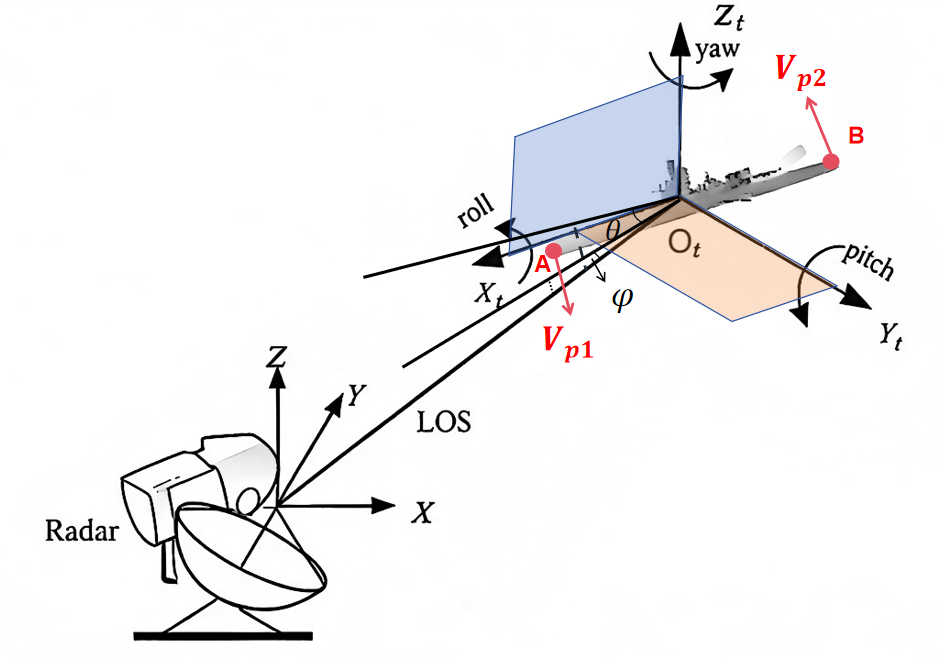}}
\caption{The spatial relationship between the radar and the target\cite{b5}}
\label{fig2}
\vspace{-1em}
\end{figure}

The spatial relationship between the radar and the target is shown in Fig.\ref{fig2}. $OXYZ$ is the radar observation coordinate system, and $O_tX_tY_tZ_t$ is the target's body-fixed coordinate system.In the initial state, the planes $OXY$ and $O_tX_tY_t$ are parallel. The line connecting the origins of the two coordinate systems, $OO_t$, represents the line-of-sight ($LOS$) direction of the radar observing the target. Its length is the radar-to-target distance $R_0$. The angle between $OO_t$ and the plane $O_tX_tZ_t$ (blue plane) is the azimuth angle $\theta$ of the observed target, while the angle between $OO_t$ and the plane $O_tX_tY_t$ (orange plane) is the elevation angle $\varphi$ of the observed target. The target attitude angle $\gamma$, induced by roll, is defined as the angle between the axis $O_tY_t$ and the plane $OXY$.

Micro-motions in different dimensions generate corresponding micro-motion velocities, and the micro-motion velocities of different parts of the target vary.Under radar observation, this manifests as a small incremental component $\Delta v$ superimposed on the overall translational velocity of each scattering center. This increment represents the projection of the total micro-motion velocity onto the radar radial direction, which differs for each scattering center.

\begin{figure}[t]
    \centering
    \subfloat[]{\includegraphics[width=0.22\textwidth]{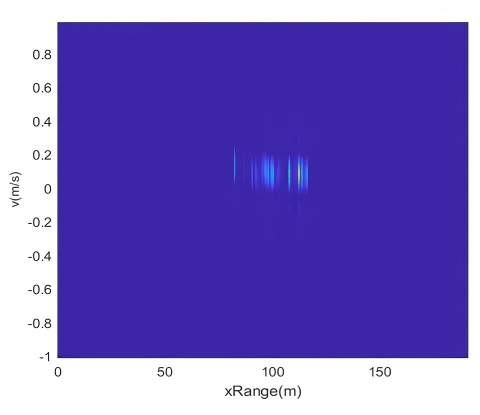}%
    \label{fig:3a}}
    \hfil
    \subfloat[]{\includegraphics[width=0.235\textwidth]{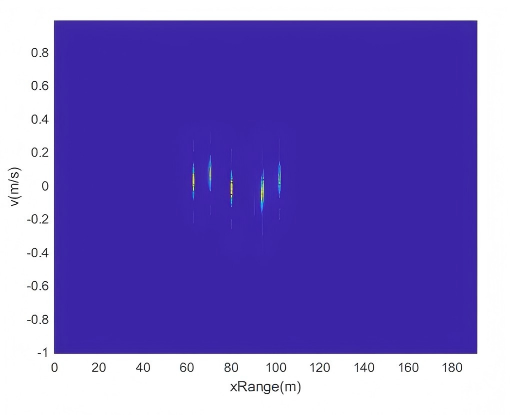}%
    \label{fig:3b}}
    \caption{Rigid/Non-rigid target RV Map: (a) ship;\\(b) corner reflector array}
    \label{fig:3}
 \vspace{-1em}
\end{figure}

Evidently, ship is rigid body, while corner reflector array is not. For rigid target, there exists a well-defined kinematic con-straint relationship between the micro-motion velocities of scattering centers on it. Taking the pitch motion of the ship in Fig. 2 as an example: two scattering centers are selected at the bow (A) and the stern (B), respectively.When A experiences a wave-induced velocity $V_{p1}$ as shown in the figure, the rigid body constraints of the entire hull cause B to simultaneously exhibit a corresponding velocity $V_{p2}$.In contrast, for non-rigid corner reflector arrays, since each corner reflector operates as an independent unit, the micro-motion states of scattering centers across individual reflectors exhibit random characteristics.In the target's RV Map, this is reflected as follows: the range-velocity distribution of scattering points on a rigid ship exhibits a strong linear trend, as shown in Fig. \ref{fig:3a}, whereas the corner reflector array demonstrates a more random and scattered distribution with poorer linearity, as illustrated in Fig. \ref{fig:3b}.

Since radar can only detect radial distance and velocity, for the same target in an identical scenario, the observed range and velocity spread of the target will vary with the radar's viewing angle.For rigid target (ship), its overall range and velocity distributions exhibit a coupled relationship: when the radar-detected range spread of the target is large, its velocity spread is small, and vice versa.But for non-rigid target(corner reflector array), the coupling between these two distributions is significantly weaker: there is no strong correlation between the range spread and velocity spread—they can easily exhibit either large or small spreads simultaneously.

\subsection{Feature Extraction and Recognition Method Design}
Firstly, the range-velocity coordinates $(r_i, v_i)$ and the weight $w_i$ (calculated based on scattering intensity) of each scattering center on each RV diagram are extracted, where $i\in\{1,...,N_{sc}\}$, $N_{sc}$ is the number of coordinates of target scattering centers.The following two feature computation methods are designed based on the analysis in \ref{AA}:

(1)Feature 1:Mean Weighted Square Residual of Range-Velocity Linear Fit: we calculate the $r-v$ linear regression fitting for each map using the least squares method: $\hat{v}_i=ar_i+b$, both $a$ and $b$ are fixed parameters determined through the fitting calculation.The mean weighted square residual(MWR) of each map can be calculated:

\begin{equation}
    \mathrm{MWR}=\frac{1}{N_{sc}}\sum\limits_{i=1}^{N_{sc}}w_i(v_i-\hat{v}_i)^2.
\end{equation}

MWR represents the degree of linearity among the scattering points in an RV Map. Typically, the RV Map for a ship demonstrates significant linearity, corresponding to a smaller MWR value. In contrast, the pattern observed for a corner reflector array is the opposite.

(2)Feature 2:Range-Velocity Distribution Complementary Contrast Factor:we calculate the respective standard deviations to characterize the spread distribution in the range and velocity dimensions($\sigma_r$ and $\sigma_v$):
\begin{equation}
    \sigma_{r}=\sqrt{\frac{1}{N_{sc}}\sum_{i=1}^{N_{sc}}(r_{i}-\overline{r})^{2}},\sigma_{{v}}=\sqrt{\frac{1}{N_{sc}}\sum_{i=1}^{N_{sc}}(v_{i}-\overline{v})^{2}}.
\end{equation}
Then we calculate a complementary contrast factor (CCF) as:
\begin{equation}
\mathrm{CCF}=\left|\frac{\sigma_r-\sigma_v}{\sigma_r+\sigma_v+\epsilon}\right|,
\end{equation}
where $\epsilon$ is a very small number to prevent division by zero, set here as $\epsilon=10^{-6}$.When $\sigma_r$ and $\sigma_v$ are simultaneously large or small, the CCF will be relatively small, indicating non-rigid characteristics (corner reflector array). Conversely, when the two exhibit a complementary relationship (one large and one small), the CCF becomes larger, representing rigid characteristics (ship). It is important to note that since $r_{i}$ (range) and $v_{i}$(velocity) typically differ in order of magnitude, they need to be normalized to a comparable scale before calculating the CCF.
\begin{figure}[t]
    \centering
    \includegraphics[scale=0.22]{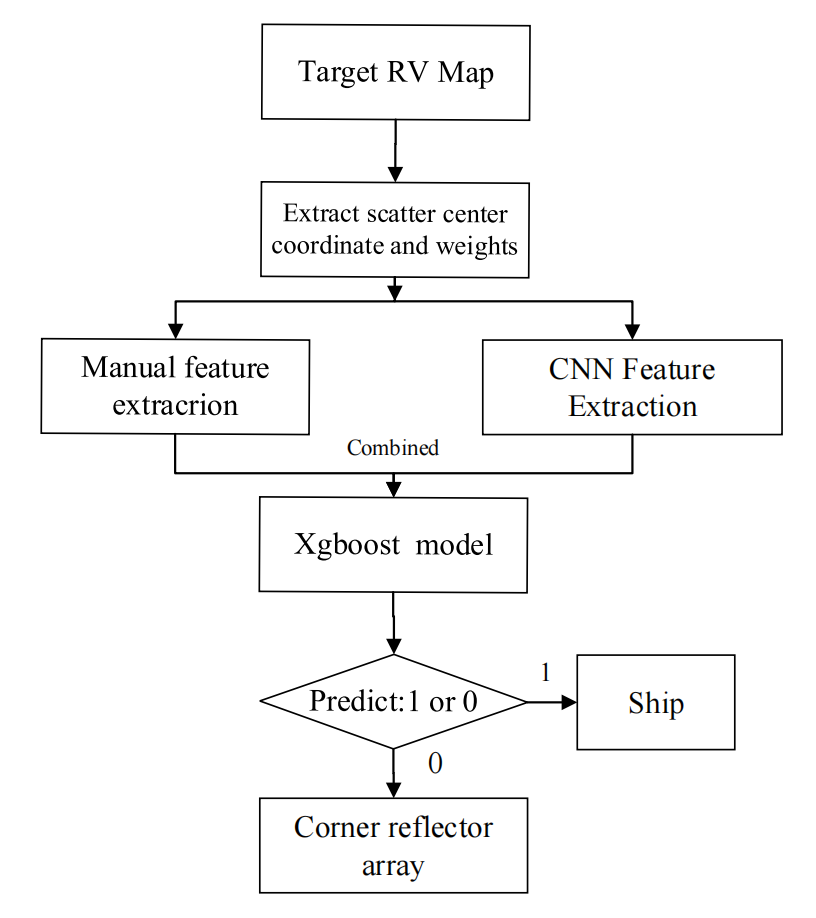}
    \caption{The flowchart of recognize corner reflector array and ship based on feature extraction from target RV Maps}
    \label{fig4}
  \vspace{-1em}
\end{figure}

 Particularly, even when normalized to the same order of magnitude, the differences between the $\sigma_r$ and $\sigma_v$ cannot be readily dismissed, thus making them viable features for recognition. In addition, we employ a CNN for automatic feature extraction(denoted as $\lambda_{CNN}$) and combine these learned features with the manually designed features as input to the recognition method\cite{b14,b15,b16}.Considering the varying importance of these features for target recognition, we have selected XGBoost as our final classifier model\cite{b17,b18,b19}.This model, through its powerful ensemble learning capabilities and built-in regularization, significantly enhances training speed and resistance to overfitting while ensuring model accuracy. We assign the label $'1'$ to ship targets and $'0'$ to corner reflector arrays. The feature vector $[\mathrm{MWR}, \mathrm{CCF}, \sigma_r, \sigma_v,\lambda_{CNN}]$ extracted from the target RV Map is then used as input to the model for supervised training\cite{b20,b21}.The model prediction output discriminates whether the target is a ship or a corner reflector array. Fig.\ref{fig4} presents the complete workflow of the recognition process for using the targets' RV Maps.
\begin{figure}[t]
    \centering
    \subfloat[]{\includegraphics[width=0.24\textwidth]{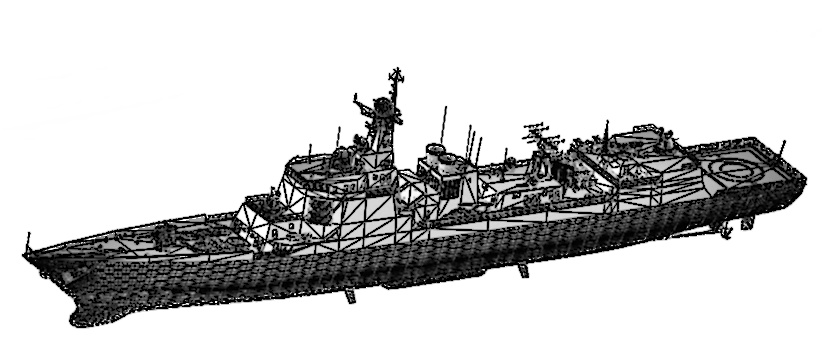}%
    \label{fig:5a}}
    \hfil
    \subfloat[]{\includegraphics[width=0.24\textwidth]{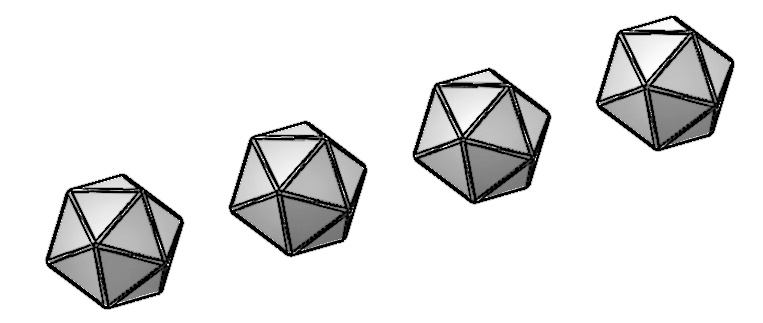}%
    \label{fig:5b}}
    \caption{Schematic of the target models:(a) ship;\\(b) corner reflector array}
    \label{fig:5}
    \vspace{-1em}
\end{figure}
\section{Experiments}

Simulated experimental data is used to validate the performance of the proposed method.The targets include one ship and one array composed of four linearly arranged icosahedral corner reflectors, as shown in Fig.\ref{fig:5}. The ship target has a length of approximately $144 m$, while each corner reflector has a diameter of about $1m$, with an inter-element spacing of roughly $40 m$. We employ DualSPHysics (an open-source SPH-based fluid/multi-physics solver) to set up irregular waves with a JONSWAP spectrum and apply a second-order (Stokes II) nonlinear correction to simulate realistic ocean wave conditions\cite{b22}.Considering that different wave amplitudes have distinct effects on the target's motion state, we configured wave scenarios with significant wave heights of $0.1m, 1m$, and $2m$ respectively. Motion simulations were conducted for both the ship and the corner reflector array under identical conditions, yielding attitude angle variation data (yaw, pitch, roll) for the targets.Based on the spatial relationship between the radar observation coordinate system and the target body coordinate system, we transformed the target's intrinsic attitude angle variations into changes in the azimuth and elevation angles observed by the radar.The frequency response of the target at different aspect angles is then calculated using CST Studio Suite(CST) software, which implicitly contains information about the target's motion state.The frequency band used here is the Ku-band.

Particularly, according to Eq.\ref{eq5},we can reformulate it into the following form:
\begin{IEEEeqnarray}{rCl}
    \hat{R}_n(f) &=& 
    \sqrt{\frac{1}{\mu}}\,
    \mathrm{rect}\!\left(\frac{f}{B}\right)
    \exp\!\left(-j2\pi \frac{f^2}{2\mu}\right)
    \nonumber\\[3pt]
    &&\times
    \sum_{k=1}^{K} \sigma_k 
    \exp(-j2\pi f_n \tau_{n,k})
    \label{eq:13}
\end{IEEEeqnarray}
where $\sum_{k=1}^{K} \sigma_k \exp(-j2\pi f_n \tau_{n,k})$ the coherent sum of the frequency responses from all scattering centers on the target, which can be substituted by the target's overall frequency response obtained through software CST simulations.

The radar waveform used in the experiment is LFM. The specific simulation parameters are listed in Table \ref{tab:radar_params}.With the target directly facing the radar (both the bow of the ship and each corner reflector aligned toward the radar) defined as the 0-degree azimuth angle and counterclockwise as the positive direction, we set up 17 observation perspectives with azimuth angles ranging from 10 to 170 degrees in 10-degree increments ($\theta=10^\circ \to 170^\circ,\theta_{step}=10^\circ$) and an elevation angle of approximately 2 degrees($\varphi\approx 2^\circ)$.Under each wave height scenario and for every observation angle, 100 RV Maps were generated for both the ship and the corner reflector array targets.

\begin{table}[t]
\centering
\caption{Radar Simulation Parameters}
\label{tab:radar_params}
\begin{tabular}{lc}
\hline
\textbf{Parameter} & \textbf{Configuration} \\
\hline
Initial carrier frequency & 16GHz \\
Carrier frequency number & 8\\
Frequency step $\Delta f$ & 25MHz\\
Pulse width & 36$\mu s$ \\
Sampling frequency & 50MHz \\
Synthetic bandwidth & 200MHz \\
PRI & 250$\mu s$\\
SCR & 20dB \\
Pulse number of a CPI & 256\\
\hline
\end{tabular}

\end{table}

\begin{figure}[t]
    \centering
    \subfloat[$\theta=80^\circ$, ship]{%
        \includegraphics[width=0.485\linewidth]{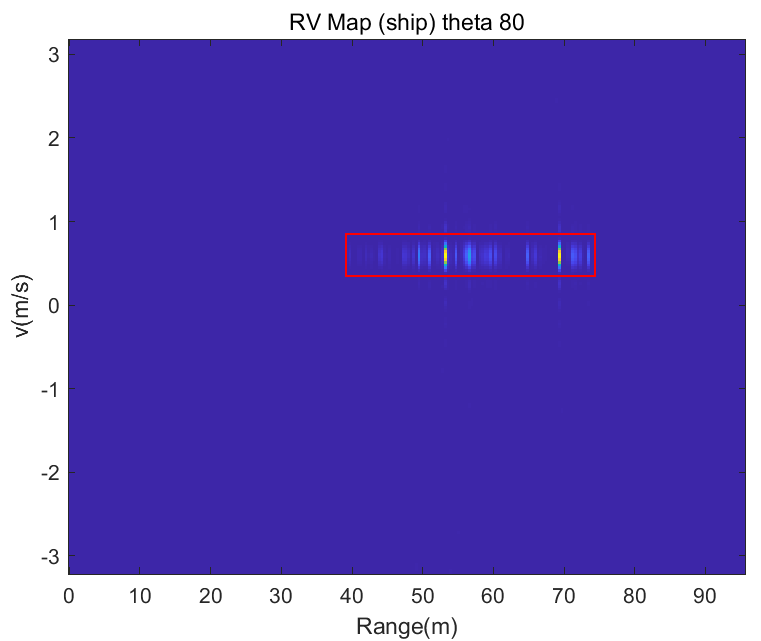}%
        \label{fig:subfig1a}
    }\hfill
    \subfloat[$\theta=80^\circ$, corner reflector array]{%
        \includegraphics[width=0.485\linewidth]{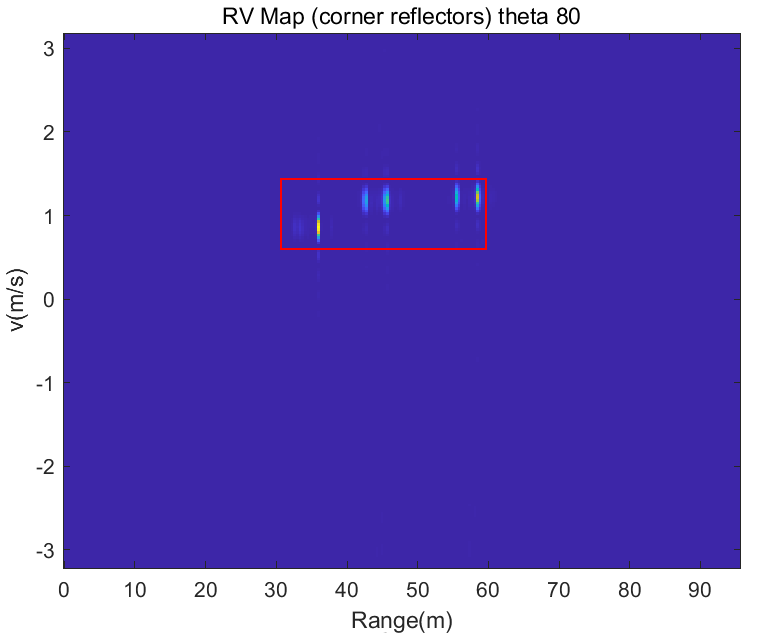}%
        \label{fig:subfig1b}
    }\\[0pt]
    \subfloat[$\theta=170^\circ$, ship]{%
        \includegraphics[width=0.485\linewidth]{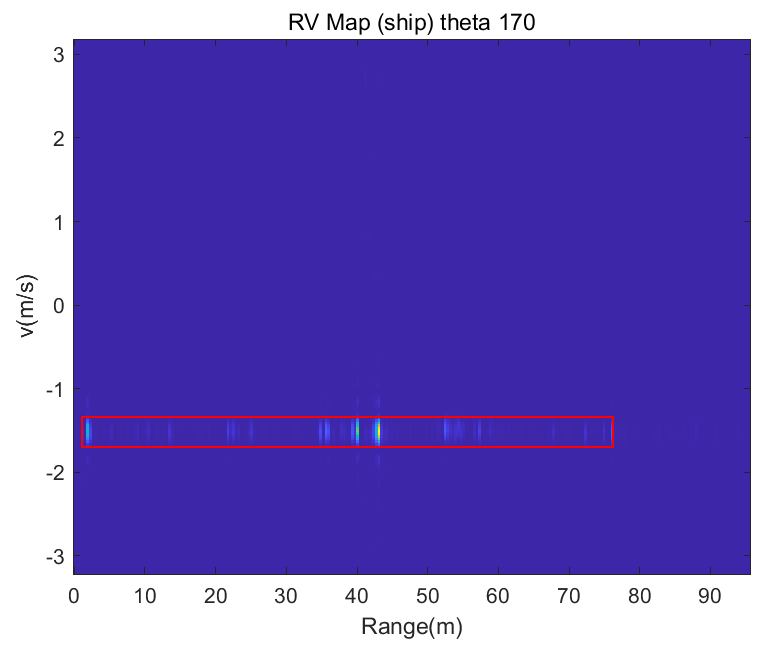}%
        \label{fig:subfig1c}
    }\hfill
    \subfloat[$\theta=170^\circ$, corner reflector array]{%
        \includegraphics[width=0.485\linewidth]{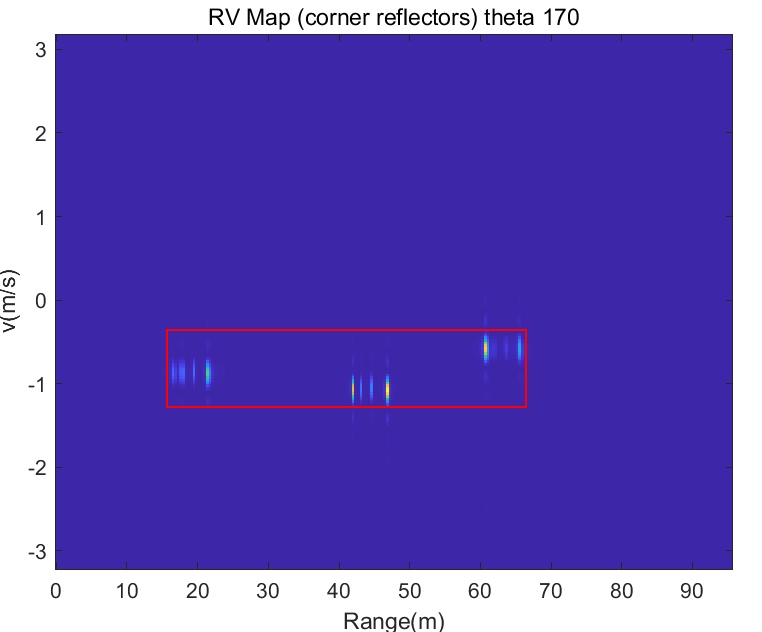}%
        \label{fig:subfig1d}
    }
    \caption{Comparison of RV Maps between ship and corner reflector array targets under two observation angles}
    \label{fig:overall}
    \vspace{-1em}
\end{figure}

Since the corner reflector array is composed of multiple independent corner reflector units, the echoes from each individual corner reflector must be linearly superimposed during simulation before signal processing to generate the target RV Map. Fig.\ref{fig:overall} demonstrates a comparison of the RV maps generated by ship and corner reflector array targets under two different observation($\theta=80^\circ, 170^\circ$) angles in a scenario with a significant wave height of $1m$. The red bounding box encloses the $(r, v)$ coordinate region of the scattering points on the target. Its horizontal extent represents the length spread of the detected target, while its vertical extent corresponds to the velocity spread. It can be visually observed that under the same observation angle, the RV Map of the ship target exhibits better linearity than that of the corner reflector array. Under different observation angles, the range spread and velocity spread of the ship target demonstrate a complementary relationship, whereas the corner reflector array does not. This observation aligns with our theoretical analysis.

\begin{figure}[t]
\centerline{\includegraphics[width=0.48\textwidth]{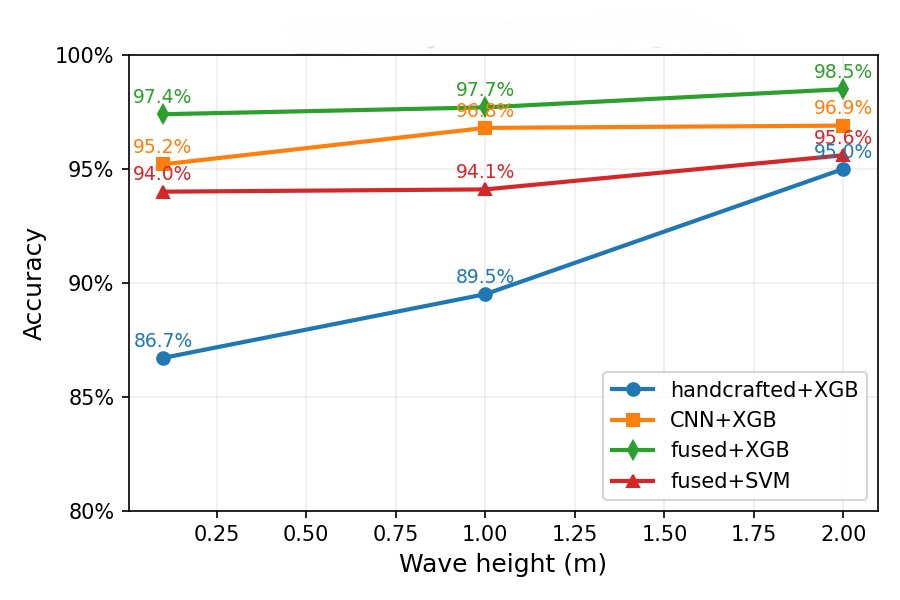}}
\caption{Variation of accuracy with significant wave height}
\label{fig6}
\vspace{-2em}
\end{figure}

The generated RV Map dataset was divided into training and testing sets in a 7:3 ratio. Training and testing were conducted using the proposed feature extraction and recognition methods. The configuration of the XGBoost model is detailed in Table \ref{xgboost}.We conducted tests on data under different wave height conditions and performed a comparative study using three experimental groups: Handcrafted Features + XGBoost, CNN Features + XGBoost, and Fused Features + XGBoost. The corresponding accuracy results are presented in Fig \ref{fig6}.As can be intuitively observed, the recognition accuracy increases with the rising wave height. This trend is primarily attributable to the fact that higher waves enhance the manifestation of non-rigid characteristics in the arrangement of the corner reflector array.Furthermore, the method we adopted demonstrates higher accuracy compared to all other control groups. This not only validates the effectiveness of integrating handcrafted features with CNN-based features to form a fused feature vector but also indicates the superior classification performance of XGBoost for this specific problem.
\begin{table}[t]
\centering
\caption{Hyperparameters of the XGBoost classifier}
\begin{tabular}{p{2cm}p{4.3cm}c}
\toprule
\textbf{Parameter} & \textbf{Description} & \textbf{Value} \\
\midrule
$n_{est}$ & Number of boosting trees & 200 \\
$depth_{max}$ & Maximum depth of each tree & 6 \\
learning rate & Step size shrinkage & 0.05 \\
subsample & Subsampling rate of training instances & 0.8 \\
colsample bytree & Fraction of features used per tree & 0.8 \\
objective & Classification objective & binary:logistic \\
eval metric & Evaluation metric & logloss \\
random state & Random seed & 42 \\
\bottomrule
\end{tabular}
\label{xgboost}
\vspace{-1em}
\end{table}

\begin{figure}[t]
    \centering
    \subfloat[]{\includegraphics[width=0.24\textwidth]{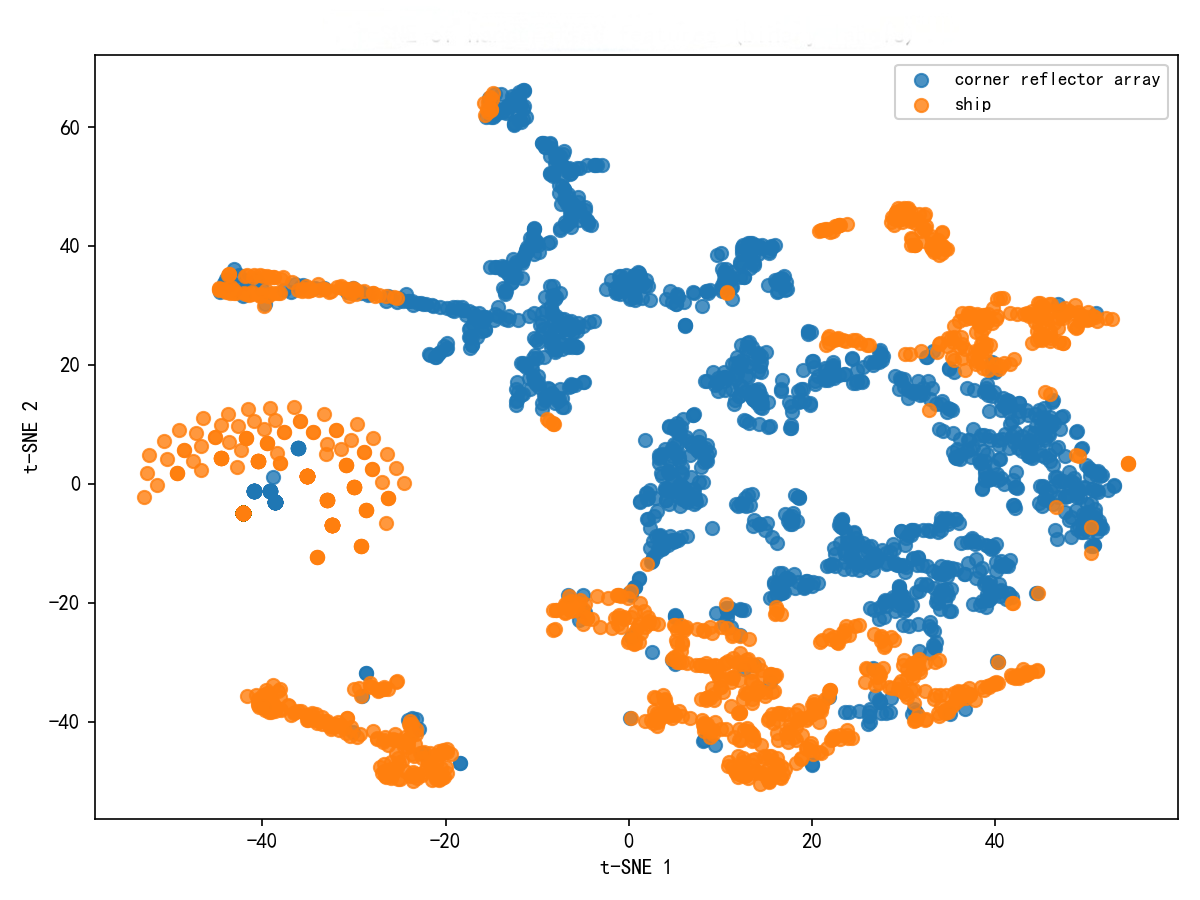}%
    \label{fig:7a}}
    \hfil
    \subfloat[]{\includegraphics[width=0.24\textwidth]{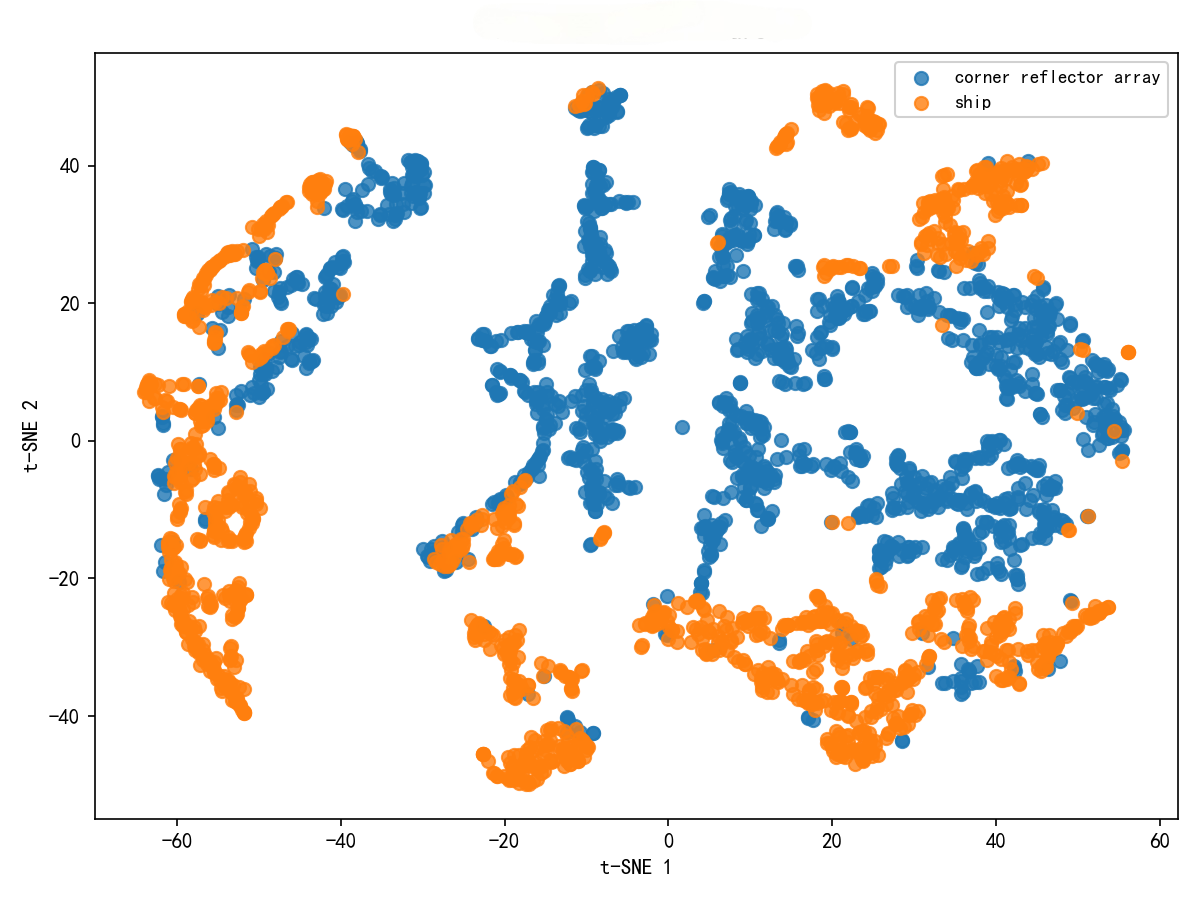}%
    \label{fig:7b}}
    \caption{T-SNE visualization of the features(blue for corner reflector array and orange for ship): (a) handcrafted features; (b) fused features}
    \label{fig:7}
    \vspace{-1em}
\end{figure}

It is particularly noteworthy that using only our custom-designed feature vectors (without incorporating CNN-extracted features) in conjunction with XGBoost achieves a recognition accuracy of over $86\%$. This result demonstrates that our feature extraction method can effectively capture the distinguishing characteristics between the two types of targets. The t-SNE visualization of the features extracted from the two types of targets is shown in Fig\ref{fig:7} (Fig \ref{fig:7a} for handcrafted features, Fig \ref{fig:7b} for fused features). As observed in Fig\ref{fig:7}, the feature vectors we extracted can effectively distinguish between the two types of targets to a considerable extent. The overlapping regions are primarily attributed to instances where the corner reflector array may exhibit rigid-body characteristics under certain scenarios, leading to misclassification. The t-SNE visualization under the fused feature set demonstrates a clearer separation between the two classes.

\section{Conclusions}
In this paper, we have developed an efficient radar target recognition framework that integrates handcrafted features with deep learning representations to discriminate between ships and corner reflector array jamming. The key innovation is our approach to leveraging the distinct multidimensional micro-motion characteristics of both targets. We design and extract two new features, MWR and CCF, directly from the target's RV Map to capture discriminative information on structural rigidity and motion distribution.These interpretable handcrafted features are combined with deep representations extracted by a CNN and fed into an XGBoost classifier. Extensive simulations under various sea states and radar viewing angles demonstrate that the proposed method achieves high recognition accuracy, particularly in higher sea conditions where non-rigid micro-motion patterns are more pronounced. It is worth emphasizing that the handcrafted multidimensional micro-motion features alone achieve over $86\%$ accuracy, highlighting both their strong interpretability and effectiveness. Future work will expand the application of our method to more diverse scenarios and evaluate the proposed multidimensional micro-motion features using real-world measured data.

\end{document}